\documentclass[10pt,twocolumn,letterpaper]{article}

\usepackage{cvpr}              

\usepackage{graphicx}
\usepackage{amsmath}
\usepackage{amssymb}
\usepackage{booktabs}
\usepackage{multirow}
\usepackage{algorithmic} 
\usepackage[table]{xcolor}
\usepackage{color}

\usepackage[pagebackref,breaklinks,colorlinks]{hyperref}

\usepackage[capitalize]{cleveref}
\crefname{section}{Sec.}{Secs.}
\Crefname{section}{Section}{Sections}
\Crefname{table}{Table}{Tables}
\crefname{table}{Tab.}{Tabs.}

\def\cvprPaperID{5084} 
\def\confName{CVPR}
\def\confYear{2022}

\begin{document}


\title{COTS: Collaborative Two-Stream Vision-Language Pre-Training Model \\for Cross-Modal Retrieval}

\author{Haoyu Lu$^{1,2}$~~~Nanyi Fei$^1$~~~Yuqi Huo$^1$~~~Yizhao Gao$^1$~~~Zhiwu Lu$^{1,2,}$\thanks{The corresponding author.}~~~Ji-Rong Wen$^{1,2}$\\
$^1$Gaoling School of Artificial Intelligence, Renmin University of China, Beijing, China\\
$^2$Beijing Key Laboratory of Big Data Management and Analysis Methods\\
{\tt\small \{lhy1998,~feinanyi,~bnhony,~gaoyizhao,~luzhiwu,~jrwen\}@ruc.edu.cn}
}

\maketitle

\begin{abstract}
   Large-scale single-stream pre-training has shown dramatic performance in image-text retrieval. Regrettably, it faces low inference efficiency due to heavy attention layers. Recently, two-stream methods like CLIP and ALIGN with high inference efficiency have also shown promising performance, however, they only consider instance-level alignment between the two streams (thus there is still room for improvement). To overcome these limitations, we propose a novel \textbf{CO}llaborative \textbf{T}wo-\textbf{S}tream vision-language pre-training model termed \textbf{COTS} for image-text retrieval by enhancing cross-modal interaction. In addition to instance-level alignment via momentum contrastive learning, we leverage two extra levels of cross-modal interactions in our COTS: (1) Token-level interaction -- a masked vision-language modeling (MVLM) learning objective is devised without using a cross-stream network module, where variational autoencoder is imposed on the visual encoder to generate visual tokens for each image. (2) Task-level interaction -- a KL-alignment learning objective is devised between text-to-image and image-to-text retrieval tasks, where the probability distribution per task is computed with the negative queues in momentum contrastive learning. Under a fair comparison setting, our COTS achieves the highest performance among all two-stream methods and comparable performance (but with 10,800$\times$ faster in inference) w.r.t. the latest single-stream methods. Importantly, our COTS is also applicable to text-to-video retrieval, yielding new state-of-the-art on the widely-used MSR-VTT dataset. 
\end{abstract}

\vspace{-0.3cm}
\section{Introduction}
\label{sec:intro}
\vspace{-0.1cm}

The pretrain-then-finetune paradigm has achieved great success in the field of natural language processing (NLP), where models are first pre-trained with large-scale data (e.g., BERT~\cite{devlin2018bert}, RoBERTa~\cite{liu2019roberta}, and GPT3~\cite{brown2020language}) and then finetuned for each downstream task. Recently, this practice has also shown its effectiveness in the vision-language (VL) domain~\cite{li2020oscar, chen2020uniter, zhang2021vinvl, kim2021vilt, radford2021learning, jia2021scaling, huo2021wenlan}, where the performance on various VL tasks (e.g., image-text retrieval, video-text retrieval, and visual question answering) has been significantly improved by vision-language pre-training (VLP). VLP models typically take huge image-text pairs as input and aim to learn joint image-text representations with single- and cross-modal pre-training objectives, such as masked token prediction and image-text matching.

\begin{figure*}[t]
    \centering
    \includegraphics[width=0.94\textwidth]{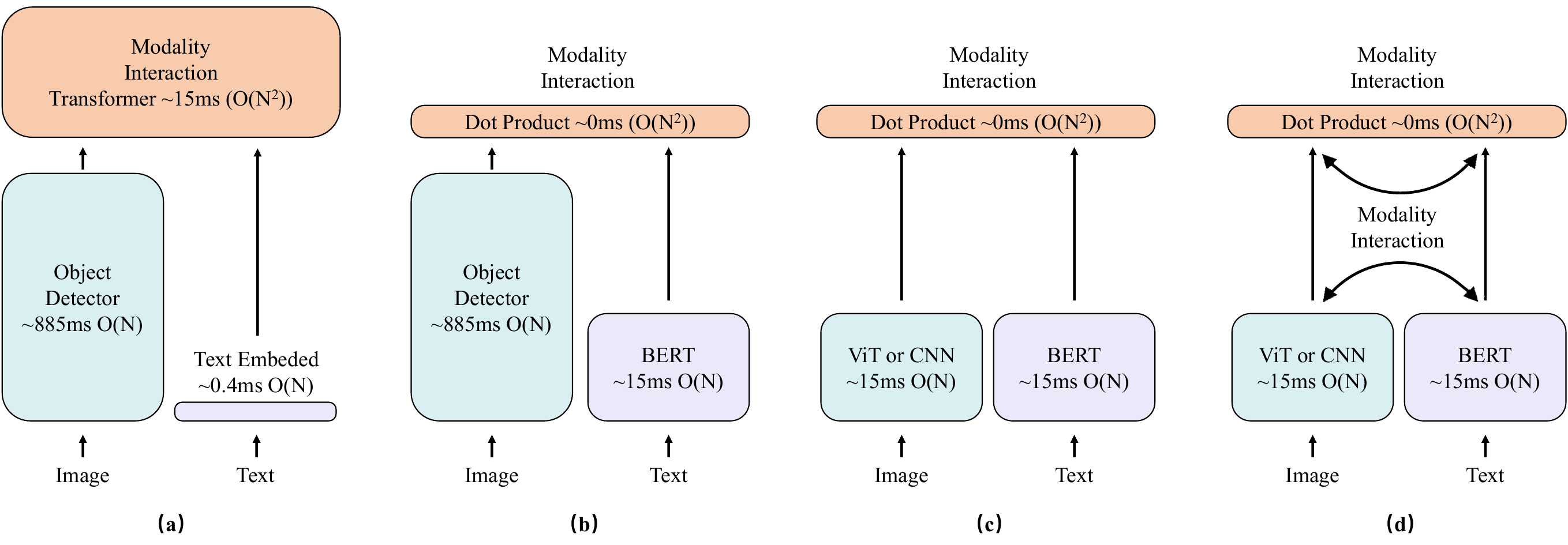}
    \vspace{-0.12in}
    \caption{
    Four categories of vision-language pre-training (VLP) models. (a) Single-stream models (e.g., Oscar~\cite{li2020oscar} and VinVL~\cite{zhang2021vinvl}). (b) Two-stream models with the object detector (e.g.,  LigntingDot~\cite{sun2021lightningdot}). (c) Two-stream models with instance-level interaction (e.g., CLIP~\cite{radford2021learning} and ALIGN~\cite{jia2021scaling}). (d) COTS: our two-stream model with multi-level interactions. The inference time and time complexity of each module are also reported, and more details can be found in Section~\ref{sec:main_results}.
    }
    \label{fig:vlm_comp}
    \vspace{-0.1in}
\end{figure*}

Existing VLP models can be divided into two groups: single-stream models and two-stream ones. Single-stream VLP models (see Figure~\ref{fig:vlm_comp}(a)) often utilize cross-modal fusion modules (e.g., Transformer~\cite{vaswani2017attention} layers) to model fine-grained interactions between image regions and text words. Although these models achieve promising performance, they have two limitations: (1) During inference, all possible query-candidate pairs need to be fed into the fusion modules to calculate similarity scores, resulting in huge computational cost. (2) To obtain meaningful image regions, single-stream models typically adopt object detectors, which are expensive in both computation and data annotation. For example, extracting object regions from a 800$\times$1,333 image takes about 900ms for Faster R-CNN~\cite{ren2015faster}, while ViT-base~\cite{dosovitskiy2021an} only needs 15ms (i.e., 60$\times$ faster).
In contrast, two-stream VLP models~\cite{kiros2014unifying, wang2016learning} apply separate image and text encoders and match image-text pairs on the final embedding level. Although two-stream models (see Figure~\ref{fig:vlm_comp}(b)--(c)) are much more efficient than single-stream ones, they only achieve sub-optimal results due to the lack of closer image-text interactions. Therefore, a few works~\cite{wu2019learning, sun2021lightningdot} (see Figure~\ref{fig:vlm_comp}(b)) reconsider object detectors, and most recent ones (e.g., CLIP~\cite{radford2021learning}, ALIGN~\cite{jia2021scaling}, and WenLan~\cite{huo2021wenlan}) resort to extra large pre-training data crawled from the Internet. However, they still fail to model fine-grained interactions between the two modalities.

To address the inefficiency of single-stream VLP models and the lack of closer vision-language interactions of two-stream ones, we propose a novel COllaborative Two-Stream vision-language pre-training model termed COTS for cross-modal retrieval, which retains the advantage of real-time inference speed and also enhances the interactions between the two modalities (see Figure~\ref{fig:vlm_comp}(d)). Concretely, we consider three levels of cross-modal interactions in our COTS: (1) Instance-level interaction -- an image-text matching learning objective at the final embedding level (typically adopted by two-stream VLP models) is devised via momentum contrastive learning~\cite{he2020momentum}, where we maintain two sample queues (one per modality) to have large size of negative samples. (2) Token-level interaction -- a novel masked vision-language modeling (MVLM) learning objective is considered without using any cross-stream network module. To this end, we first tokenize both the image and the text for each input image-text pair, where variational autoencoder~\cite{kingma2013auto} is imposed on the visual encoder (e.g., ViT~\cite{dosovitskiy2021an}) to generate visual tokens and BERT~\cite{devlin2018bert} is adopted for the text encoder. We then perform masked visual token prediction based on the unmasked visual tokens and the feature of each image's paired text, and perform masked language token prediction similarly. (3) Task-level interaction -- a novel KL-alignment learning objective is devised between text-to-image and image-to-text retrieval tasks by minimizing the Kullback-Leibler (KL) Divergence between probability distributions of the two retrieval tasks. For each image-text pair, the probability distribution of the text-to-image retrieval task is obtained with the similarities of the chosen text and its unpaired images in the negative image queue maintained in momentum contrastive learning, and we can obtain the other distribution similarly.

As the scale of pre-training data becomes large (e.g., tens of millions or even billions of image-text pairs crawled from the Internet), it is impossible to perform human-annotation and thus there inevitably exist noises in the large-scale data. Noisy data such as mis-matched image-text pairs and totally meaningless ones could bring negative effect for pre-training. In this paper, we thus propose an adaptive momentum filter (AMF) module for our COTS, which can make full use of the momentum mechanism in our contrastive learning-based training algorithm. Specifically, we first calculate the similarity scores of all image-text pairs from the dynamically maintained image and text queues to obtain an extra queue. Further, we model this queue of similarity scores as a normal distribution and filter out the noisy data with the distribution mean and variance on the fly.

Our contributions are summarized as follows:
(1) We propose a novel COllaborative Two-Stream (COTS) VLP model to improve the performance of two-stream models and retain their efficiency advantage at the same time. We achieve this by leveraging two extra levels of cross-modal interactions in addition to the typical instance-level alignment: a masked vision-language modeling (MVLM) learning objective for token-level interaction, and a KL-alignment learning objective for task-level interaction.
(2)~To alleviate the negative effect caused by the noises in large-scale pre-training data, we propose an adaptive momentum filter (AMF) module. AMF makes full use of the momentum mechanism in our instance-level alignment and adaptively filters noisy image-text pairs during pre-training.
(3) Under a fair comparison setting, our COTS achieves the highest performance among all two-stream methods and performs comparably (but 10,800$\times$ faster in inference) with the latest single-stream ones. Importantly, our COTS is also applicable to text-to-video retrieval, yielding new state-of-the-art on the widely-used MSR-VTT dataset.

\vspace{-0.1cm}
\section{Related Work}
\vspace{-0.1cm}

\noindent\textbf{Vision-Language Pre-Training.}~~Recently, VLP resorts to single-stream models or two-stream ones. Single-stream models~\cite{lu2019vilbert, huang2020pixel, chen2020uniter, gan2020large, li2020oscar, zhang2021vinvl} contain cross-modal fusion modules (e.g., Transformer~\cite{vaswani2017attention} layers) to model closer interactions between image regions and text words. Although single-stream models often achieve superior performance, they have several limitations in real-world scenarios: (1)~When performing cross-modal retrieval during inference, all possible query-candidate pairs need to be fed into the fusion modules to calculate similarity scores, resulting in huge computational cost. (2) To obtain meaningful image regions, single-stream models often adopt object detectors, which are expensive in both computation and data annotation. In contrast, two-stream models project the two modalities into a joint embedding space and align them on the final embedding level. Early two-stream models~\cite{yan2015deep, wang2016learning} only achieve sub-optimal performance because they do not consider fine-grained cross-modal interactions. More recent works (e.g., CLIP~\cite{radford2021learning}, ALIGN~\cite{jia2021scaling}, and WenLan~\cite{huo2021wenlan}) choose to improve their performance by leveraging extra large web data. However, they fail to model fine-grained interactions between the two modalities. Although the latest two-stream model LightingDot~\cite{sun2021lightningdot} considers token-level interaction, it still relies on an object detector, thus suffering from heavy computation. In this work, our COTS integrates the advantages of single-stream and two-stream models by still utilizing the two-stream architecture but enhancing the modeling of cross-modal interactions.

\noindent\textbf{Masked Vision Modeling.}~~Many previous works on VLP~\cite{chen2020uniter, li2020oscar} adopt masked vision modeling based on object tags to achieve better performance. They typically deploy a bottom-up attention mechanism~\cite{anderson2018bottom} implemented by first extracting the object tags with Faster R-CNN~\cite{ren2015faster} and then predicting the masked tags with other unmasked tags and text tokens. Although higher performance can be achieved, they commonly face two issues: (1) A heavy detector is needed to extract object tags, which is computationally expensive. For example, a Faster R-CNN detector takes 900ms to extract fine-grained region information from an image, which is nearly 60$\times$ slower than our ViT-base backbone (15ms). (2) These VLP models are not end-to-end trained, which may fail to cope with unknown objects.
The latest work~\cite{kim2021vilt} shows that simply predicting masked raw image pixels is hard to improve the performance. Different from these works, our COTS employs a variational autoencoder~\cite{kingma2013auto} as an image tokenizer to tokenize a raw image into discrete image tokens for masked vision modeling, inspired by the vision Transformer BEIT~\cite{bao2021beit}. The tokenizer is pre-trained in an end-to-end unsupervised training style, avoiding inducing handcrafted tags or heavy object detectors. Importantly, compared with predicting raw pixels directly, our choice of predicting masked image tokens is more meaningful as each image token contains specific high-level visual information. Overall, by combining masked vision modeling with masked language modeling, we devise a novel masked vision-language modeling (MVLM) objective for closer token-level interaction.

\section{Methodology}

\subsection{Framework Overview}

\begin{figure*}[t]
    \centering
    \includegraphics[width=0.96\textwidth]{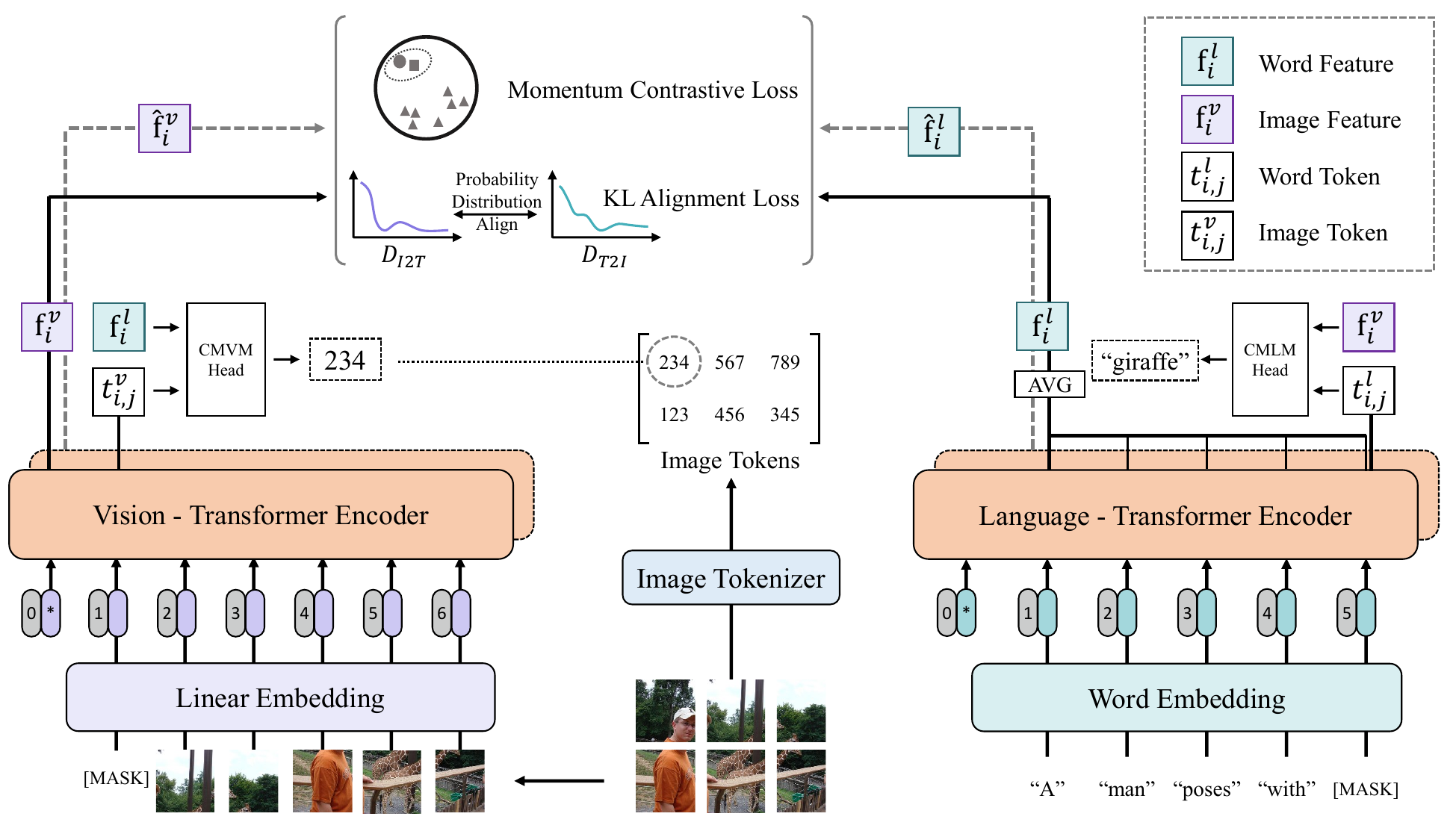}
    \vspace{-0.1in}
    \caption{
    A schematic illustration of the proposed COTS for cross-modal retrieval.
    }
    \label{fig:cross-modal}
    \vspace{-0.1in}
\end{figure*}

The goal of our COTS model for VLP is to learn two separate encoders that can embed image and text samples into the same semantic space for effective cross-modal retrieval. As illustrated in Figure~\ref{fig:cross-modal}, images and texts are encoded by the vision Transformer and the language Transformer, respectively. We then devise three levels of cross-modal interactions as the pre-training objectives of our COTS. Concretely, the instance-level interaction aligns the global features of paired images and texts by momentum cross-modal contrastive learning, which is inspired by the single-modal MoCo~\cite{he2020momentum}. To model closer interactions than instance-level alignment, we propose to devise a masked vision-language modeling (MVLM) loss to enhance token-level interaction. MVLM has two parts: cross-modal masked vision modeling (CMVM) and cross-modal masked language modeling (CMLM). For each image, CMVM aims to predict the label of the masked image patch token based on unmasked ones together with the global feature of its paired text. CMLM does similarly on the language side. Further, we consider task-level interaction in our COTS, which aims to align the probability distributions of text-to-image and image-to-text retrieval tasks. In addition, to cope with the noises in the large-scale pre-training data, we propose an adaptive momentum filter (AMF) module, which is seamlessly integrated into the pre-training process.

Our choice of adopting the two-stream architecture in COTS has two main advantages: (1) Real-time inference speed -- the separate image and text encoders allow us to compute the features of candidates beforehand for cross-modal retrieval tasks, and only a simple dot product needs to be calculated for each query-candidate pair. (2) Applicability to text-to-video retrieval -- without any modification, our COTS can be directly applied to the text-to-video retrieval task, where the video representation can be obtained by averaging frame embeddings obtained by the image encoder. More details are given in Section~\ref{video}.

\subsection{Training Objectives}

\subsubsection{Token-Level Interaction}

We devise a masked vision-language modeling (MVLM) loss to enhance the token-level interaction in our COTS, which can be further split into two parts: cross-modal masked vision modeling (CMVM) and cross-modal masked language modeling (CMLM). To improve the practice~\cite{chen2020uniter, li2020oscar} of predicting masked image region tags with heavy object detectors, we introduce CMVM based on an image tokenizer inspired by BEIT~\cite{bao2021beit}. For each image, the objective of CMVM is to predict the labels of masked image tokens with the unmasked image patches and paired text.

Formally, let $\mathcal{D} = \{(v_i, l_i)\}_{i=1}^{N}$ denote the training dataset, where $(v_i, l_i)$ is the $i$-th image-text pair. For each raw image $v_i$, we first utilize the pre-trained discrete variational auto-encoder (dVAE)~\cite{ramesh2021zero} as the image tokenizer to obtain a sequence of $24\times24$ discrete image tokens $\mathcal{T}^v_i = \{ t^v_{i,j} \in \mathcal{V}^v \}_{j=1}^{576}$, where $t^v_{i,j}$ is the $j$-th token of image $v_i$ and $\mathcal{V}^v$ is the vocabulary of discrete image tokens. Meanwhile, the raw image is split into $24\times24$ patches, which are fed into a vision Transformer~\cite{dosovitskiy2021an} to obtain their embeddings. We then predict the label of each masked token based on the summation of the masked token embedding (which is already fused with unmasked token embeddings) and the global embedding of the paired text. The CMVM loss can thus be formulated as:
\begin{equation}
\mathcal{L}_{\text{CMVM}} = - \mathbb{E}_{(v_i, l_i) \sim \mathcal{D}} \log P(t^v_{i,j}|t^v_{i, \backslash j}, l_i),
\end{equation}
where $t^v_{i,j}$ denotes the target/masked image token, and $t^v_{i, \backslash j} = \mathcal{T}^v_i \backslash \{t^v_{i,j}\}$ denotes the unmasked image tokens.

Similar to CMVM, for each piece of text $l_i$, the objective of CMLM is to predict the label of each masked word token based on unmasked ones and the paired image:
\begin{equation}
\mathcal{L}_{\text{CMLM}} = - \mathbb{E}_{(v_i, l_i) \sim  \mathcal{D}} \log P(t^l_{i,j}|t^l_{i, \backslash j}, v_i),
\end{equation}
where $t^l_{i,j}$ denotes the target/masked text word token, and $t^l_{i, \backslash j}$ denotes the unmasked ones. The total loss of our token-level cross-modal interaction is then defined as:
\begin{equation}
\mathcal{L}_{\text{token}} = \mathcal{L}_{\text{CMVM}} + \mathcal{L}_{\text{CMLM}}.
\end{equation}

\subsubsection{Instance-Level Interaction}
\label{instance}

To model the instance-level interaction of two modalities (i.e., global feature alignment) in our COTS, we adopt a cross-modal momentum contrastive learning (MCL) algorithm inspired by the single-modal MoCo~\cite{he2020momentum}, which provides a mechanism of dynamically maintaining negative sample queues for contrastive learning. Since the two queues (one for each modality) used in our MCL successfully decouple the queue size from the mini-batch size, the size of negative samples (crucial for contrastive learning) can be much larger than the mini-batch size.

Concretely, let $f^v$ (with parameters $\theta^v$) and $f^l$ (with parameters $\theta^l$) denote the image and text encoders, respectively. We adopt two extra momentum encoders $\hat{f}^v$ (with parameters $\hat{\theta}^v$) and $\hat{f}^l$ (with parameters $\hat{\theta}^l$) for the vision and language modalities, respectively. The parameters of momentum encoders are updated by:
\begin{align}
\hat{\theta}^v & = m \cdot \hat{\theta}^v + (1 - m) \cdot \theta^v, \\
\hat{\theta}^l & = m \cdot \hat{\theta}^l + (1 - m) \cdot \theta^l,
\end{align}
where $m$ is the momentum hyper-parameter.

Further, we maintain two queues $\mathcal{Q}^v = \{\mathbf{\hat{q}}^v_j\}_{j=1}^{N_q}$ and $\mathcal{Q}^l = \{\mathbf{\hat{q}}^l_j\}_{j=1}^{N_q}$, where $\mathbf{\hat{q}}^v_j$/$\mathbf{\hat{q}}^l_j$ denotes the momentum feature vector, and $N_q$ denotes the queue size. Samples in each mini-batch $\mathcal{B} =\{(v_i, l_i)\}_{i=1}^{N_b} \subseteq \mathcal{D}$ ($N_b = |\mathcal{B}| \ll N_q$) are fed into current momentum encoders to obtain their momentum feature vectors, which are then pushed into corresponding queues \emph{after loss calculation}. Meanwhile, the earliest $N_b$ momentum feature vectors in each queue are popped out. Given each image in a data batch, by regarding its paired text as the positive sample and all samples in $\mathcal{Q}^l$ as negative ones, we define the image-to-text contrastive loss as ($\tau$ is the temperature hyper-parameter):
\begin{equation}
\mathcal{L}_\text{I2T} \hspace{-1pt}=\hspace{-1pt} -\frac{1}{N_b} \hspace{-3pt}\sum_{(v_i, l_i) \in \mathcal{B}} \hspace{-7pt}\log \frac{\text{pos}(\mathbf{f}^v_i, \mathbf{\hat{f}}^l_i, \tau)}{\text{pos}(\mathbf{f}^v_i, \mathbf{\hat{f}}^l_i, \tau) \hspace{-2pt}+\hspace{-2pt} \text{neg}(\mathbf{f}^v_i, \mathcal{Q}^l, \tau)},
\label{eq:cl_i2t}
\end{equation}
where $\mathbf{f}^v_i = f^v(v_i)$, $\mathbf{\hat{f}}^l_i = \hat{f}^l(l_i)$, and
\begin{align}
\text{pos}(\mathbf{f}^v_i, \mathbf{\hat{f}}^l_i, \tau) & = \exp(\mathbf{f}^v_i \cdot \mathbf{\hat{f}}^l_i / \tau), \\
\text{neg}(\mathbf{f}^v_i, \mathcal{Q}^l, \tau) & = \sum_{\mathbf{\hat{q}}^l_j \in \mathcal{Q}^l} \exp(\mathbf{f}^v_i \cdot \mathbf{\hat{q}}^l_j / \tau).
\end{align}
The similarity of two feature vectors is measured by dot product here. Similarly, given each text in a data batch, we define the text-to-image contrastive loss as:
\begin{equation}
\mathcal{L}_\text{T2I} \hspace{-1pt}=\hspace{-1pt} -\frac{1}{N_b} \hspace{-3pt}\sum_{(v_i, l_i) \in \mathcal{B}} \hspace{-7pt}\log \frac{\text{pos}(\mathbf{f}^l_i, \mathbf{\hat{f}}^v_i, \tau)}{\text{pos}(\mathbf{f}^l_i, \mathbf{\hat{f}}^v_i, \tau) \hspace{-2pt}+\hspace{-2pt} \text{neg}(\mathbf{f}^l_i, \mathcal{Q}^v, \tau)},
\end{equation}
where $\mathbf{f}^l_i = f^l(l_i)$, and $\mathbf{\hat{f}}^v_i = \hat{f}^v(v_i)$. The total loss of our instance-level cross-modal interaction is then defined as:
\begin{equation}
\mathcal{L}_{\text{inst}} = \mathcal{L}_{\text{I2T}} + \mathcal{L}_{\text{T2I}}.
\end{equation}

\subsubsection{Task-Level Interaction}

As we can see from Eq.~(\ref{eq:cl_i2t}) that, for each image $v_i$ in a mini-batch, the image-to-text contrastive objective is actually maximizing the probability of matching its paired text $l_i$ against the unmatched samples in $\mathcal{Q}^l$ (so does the text side). That is, the instance-level feature alignment only cares about maximizing one particular probability in the whole probability distribution of the image-to-text/text-to-image retrieval task, and fails to capture a higher level interaction between two modalities. To fill the void in the literature, we propose to align the probability distributions of two cross-modal retrieval tasks as our task-level interaction.

Concretely, for each image-text pair $(v_i, l_i) \in \mathcal{B}$, we define the probability distribution of the image-to-text task as:
\begin{equation}
\mathcal{D}_{\text{I2T}} = [p(\mathbf{f}^v_i, \mathbf{\hat{f}}^l_i), p(\mathbf{f}^v_i, \mathbf{\hat{q}}^l_1), \cdots, p(\mathbf{f}^v_i, \mathbf{\hat{q}}^l_{N_q})],
\end{equation}
where
\begin{equation}
p(\mathbf{f}^v_i, \mathbf{\hat{f}}^l_i) = \frac{\exp(\mathbf{f}^v_i \cdot \mathbf{\hat{f}}^l_i / \tau)}{\sum_{\mathbf{\hat{f}} \in \{\mathbf{\hat{f}}^l_i\} \cup \mathcal{Q}^l} \exp(\mathbf{f}^v_i \cdot \mathbf{\hat{f}} / \tau)},
\end{equation}
and $p(\mathbf{f}^v_i, \mathbf{\hat{q}}^l_j)$ ($\mathbf{\hat{q}}^l_j \in \mathcal{Q}^l$, $j=1,2,\cdots,N_q$) can be calculated in the same way. Similarly, we obtain the probability distribution of the text-to-image task as:
\begin{equation}
\mathcal{D}_{\text{T2I}} = [p(\mathbf{f}^l_i, \mathbf{\hat{f}}^v_i), p(\mathbf{f}^l_i, \mathbf{\hat{q}}^v_1), \cdots, p(\mathbf{f}^l_i, \mathbf{\hat{q}}^v_{N_q})].
\end{equation}
The learning objective of our task-level cross-modal interaction is then formulated as minimizing the symmetric Kullback-Leibler (KL) Divergence between $\mathcal{D}_{\text{I2T}}$ and $\mathcal{D}_{\text{T2I}}$:
\begin{equation}
\mathcal{L}_{\text{task}} \hspace{-1pt}=\hspace{-1pt} \frac{1}{N_b} \hspace{-3pt}\sum_{(v_i, l_i) \in \mathcal{B}} \hspace{-8pt}( \text{KL}(\mathcal{D}_{\text{I2T}} || \mathcal{D}_{\text{T2I}}) \hspace{-1pt}+\hspace{-1pt} \text{KL}(\mathcal{D}_{\text{T2I}} || \mathcal{D}_{\text{I2T}}) ).
\end{equation}

\subsection{Adaptive Momentum Filter}

Large-scale web-crawled data inevitably contain noises, which could bring negative effect for pre-training. Therefore, based on the momentum mechanism adopted in our COTS, we propose an adaptive momentum filter (AMF) module to adaptively filter noisy image-text pairs.

As introduced in the instance-level interaction, our COTS dynamically maintains two sample queues $\mathcal{Q}^v$ and $\mathcal{Q}^l$ for momentum contrastive learning. Since paired images and texts are pushed into or popped out of the corresponding queue simultaneously, $\mathbf{\hat{q}}^v_j \in \mathcal{Q}^v$ and $\mathbf{\hat{q}}^l_j \in \mathcal{Q}^l$ ($j=1,2,\cdots,N_q$) are also paired. We can then calculate a similarity score for each pair $(\mathbf{\hat{q}}^v_j, \mathbf{\hat{q}}^l_j)$ by dot product. In this way, we obtain an extra similarity queue $\mathcal{Q}^s = \{ \mathbf{\hat{q}}^v_j \cdot \mathbf{\hat{q}}^l_j | \mathbf{\hat{q}}^v_j \in \mathcal{Q}^v, \mathbf{\hat{q}}^l_j \in \mathcal{Q}^l \}_{j=1}^{N_q}$, which is also dynamically maintained along with the two sample queues.

Note that the similarity queue $\mathcal{Q}^s$ can be seen as a sampling of the similarity score distribution at the current training iteration. We first calculate its mean $\mu$ and standard deviation $\sigma$ as the estimations of those of the similarity score distribution. We then obtain the threshold value $s_{\text{AMF}}$ based on $\mu$ and $\sigma$ (e.g., $s_{\text{AMF}} = \mu - 2\sigma$) for our AMF. Finally, we use this threshold to filter the current data batch $\mathcal{B}$ before we compute the losses:
\begin{equation}
\mathcal{B}^* = \{ (v_i, l_i) | \mathbf{\hat{f}}^v_i \cdot \mathbf{\hat{f}}^l_i > s_{\text{AMF}}, (v_i, l_i) \in \mathcal{B} \}.
\end{equation}
In this work, $s_{\text{AMF}}$ changes in different training iterations as the similarity queue is changing. Specifically, when AMF is adopted in our full COTS, we use $\mathcal{B}^*$ instead of $\mathcal{B}$ in each iteration for loss computation, but we still push all samples in $\mathcal{B}$ into $\mathcal{Q}^v$ and $\mathcal{Q}^l$ after loss computation.

\begin{table*}[t]
    \centering
    \scalebox{0.90}{
    \tabcolsep5.0pt
    \begin{tabular}{lccccccccccccc}
    \toprule
    \multirow{3}{*}{Model} & \multirow{3}{*}{\# PT Pairs} &  \multicolumn{6}{c}{Flickr30K (1K)} & \multicolumn{6}{c}{MSCOCO (5K)} \\
    & &  \multicolumn{3}{c}{I2T Retrieval} & \multicolumn{3}{c}{T2I Retrieval} & \multicolumn{3}{c}{I2T Retrieval} & \multicolumn{3}{c}{T2I Retrieval} \\
    & & R@1 & R@5 & R@10 & R@1 & R@5 & R@10 & R@1 & R@5 & R@10 & R@1 & R@5 & R@10 \\ \midrule
    \bf{Single-Stream:} &   \\
    ViLBERT-Base~\cite{lu2019vilbert} & 3.1M & - & - & - & 58.2 & 84.9 & 91.5 & - & - & - &  - & - & - \\
    Pixel-BERT-R50~\cite{huang2020pixel} & 5.6M & 75.7 & 94.7 & 97.1 & 53.4 & 80.4 & 88.5 & 59.8 & 85.5 & 91.6 &  41.1 & 69.7 & 80.5 \\
    Pixel-BERT-X152~\cite{huang2020pixel} & 5.6M & \bf87.0 & \bf98.9 & \bf99.5 & 71.5 & 92.1 & 95.8 & 63.6 & 87.5 & 93.6 &  50.1 & 77.6 & 86.2 \\
    Unicoder-VL~\cite{li2020unicoder} & 3.8M & 86.2 & 96.3 & 99.0 & 71.5 & 91.2 & 95.2 & 62.3 & 87.1 & 92.8 &  48.4 & 76.7 & 85.9 \\
    UNITER-Base~\cite{chen2020uniter} & 9.6M & 85.9 & 97.1 & 98.8 &  72.5 & 92.4 & \bf96.1 & 64.4 & 87.4 & 93.1 &50.3 & 78.5 & 87.2 \\
    ERNIE-ViL-base~\cite{yu2021ernie} & 3.8M &  86.7 & 97.8 & 99.0 & 74.4 & 92.7 & 95.9 & - & - & - & - & - & - \\
    VILLA-Base~\cite{gan2020large} & 9.6M & 86.6 & 97.9 & 99.2 & \bf74.7 & \bf92.9 & 95.8 & - & - & - & - & - & - \\
    OSCAR-Base~\cite{li2020oscar} & 6.5M & - & - & - & - & - & - & 70.0 & 91.1 & 95.5 & 54.0 & 80.8 & 88.5 \\ 
    ViLT~\cite{kim2021vilt} & 9.9M & 83.5 & 96.7 & 98.6 & 64.4 & 88.7 & 93.8 & 61.5 & 86.3 & 92.7 &  42.7 & 72.9 & 83.1 \\ 
    VinVL-Base~\cite{zhang2021vinvl} & 8.9M & - & - & - & - & - & -  & \bf74.6 & \bf92.6 & \bf96.3 & \bf58.1 & \bf83.2 & \bf90.1 \\
    \midrule
    \bf{Two-Stream:}  & \\
    {\color{gray}VSE$\infty^{*\dagger}$~\cite{chen2021learning}} & {\color{gray}-}  & {\color{gray}88.7} & {\color{gray}98.9} & {\color{gray}99.8} & {\color{gray}76.1} & {\color{gray}94.5} & {\color{gray}97.1} & {\color{gray}68.1} & {\color{gray}90.2} & {\color{gray}95.2} &  {\color{gray}52.7} & {\color{gray}80.2} & {\color{gray}88.3} \\
    {\color{gray}COOKIE$^{*\dagger}$~\cite{wen2021cookie}} & {\color{gray}5.9M} &{\color{gray}89.0} & {\color{gray}98.9} & {\color{gray}99.7} & {\color{gray}75.6} & {\color{gray}94.6} & {\color{gray}97.2} & {\color{gray}71.6} & {\color{gray}90.9} & {\color{gray}95.4} & {\color{gray}54.5} & {\color{gray}81.0} & {\color{gray}88.2} \\
    Frozen in time~\cite{bain2021frozen} & 5.5M & - & - & - & 61.0 & 87.5 & 92.7 & - & - & - &  - & - & - \\
    LightningDOT~\cite{sun2021lightningdot} & 9.5M & 83.9 & 97.2 & 98.6 & 69.9 & 91.1 & 95.2 & 60.1 & 85.1 & 91.8 &  45.8 & 74.6 & 83.8\\
    COOKIE~\cite{wen2021cookie} & 5.9M  &84.7 & 96.9 & 98.3 & 68.3 & 91.1 & 95.2 & 61.7 & 86.7 & 92.3 & 46.6 & 75.2 & 84.1 \\
    COTS~(ours) & 5.3M  & 88.2 & 98.5 & 99.7 & 75.2 & 93.6 & 96.5 & 66.9 & 88.8 & 94.0 &  50.5 & 77.6 & 86.1 \\
    COTS~(ours) & 15.3M & 90.6 & 98.7 & 99.7 & 76.5 & 93.9 & 96.6  & 69.0 & 90.4 & 94.9 & 52.4 & 79.0 & 86.9 \\ 
    COTS$^\dagger$~(ours) & 15.3M  & \bf91.7 & \bf99.0 & \bf99.9 & \bf78.3 & \bf94.9 & \bf97.2  & \bf70.6 & \bf91.0 & \bf95.3 & \bf53.7 & \bf80.2 & \bf87.8 \\ 
    \bottomrule
    \end{tabular}}
    \vspace{-0.1in}
    \caption{Comparative results for image-text retrieval on the Flickr30K (1K) test set and MSCOCO (5K) test set. Notations: \# PT Pairs -- the number of image-text pairs for pre-training; I2T Retrieval -- image-to-text retrieval; T2I Retrieval -- text-to-image retrieval. $^\dagger$ Ensemble results of two models. $^*$ Models that utilize {\color{gray}940M tagged images} for visual encoder pre-training. 
    }  
    \label{tab:image_sota}
    \vspace{-0.1in}
\end{table*}

\section{Experiments}

\subsection{Datasets and Settings}

\noindent\textbf{Pre-Training Datasets.}~~We use two image-text datasets for pre-training our COTS: (1) \textbf{CC4M} contains 4 million images and 5.3 million captions from Conceptual Captions~(CC3M)~\cite{sharma2018conceptual}, SBU~\cite{ordonez2011im2text}, VG~\cite{krishna2017visual}, MSCOCO ~\cite{lin2014microsoft} and Flickr30K~\cite{plummer2015flickr30k}. (2)~\textbf{CC14M} consists of CC4M and CC12M~\cite{changpinyo2021conceptual} (about 2 million urls are now invalid), which contains 14 million images and 15.3 million captions in total. Note that CC14M is much noisier than CC4M.

\noindent\textbf{Downstream Datasets.}~~We make downstream evaluation of our COTS on three widely-used benchmark datasets: (1)~\textbf{MSCOCO}~\cite{lin2014microsoft} is a large image-text dataset of 123,287 images, where each image is annotated with 5 captions. As in~\cite{kim2021vilt}, we adopt the Karpathy split of MSCOCO: 5,000 images for testing, another 5,000 for validation, and the rest 113,287 images for training. (2)~\textbf{Flickr30K}~\cite{plummer2015flickr30k} contains 31,000 images and 158,915 captions totally. Each image is often annotated with 5 captions. Following the split in \cite{frome2013devise}, we use 1,000 images for testing, another 1,000 for validation, and the rest for training. (3)~To show the general applicability of our COTS, we also conduct experiments on a video-text dataset \textbf{MSR-VTT}~\cite{xu2016msr}, which has 10K YouTube videos and 200K captions. As in~\cite{yang2021taco}, we report our results under both the 1KA and 7K splits.

\noindent\textbf{Text and Image Encoders.}~~In our COTS, we follow~\cite{sun2021lightningdot} and adopt a BERT-base~\cite{devlin2018bert} model as our text encoder, which contains a total of 12 Transformer layers with 768 hidden units and 12 heads. Further, for computation efficiency, we use ViT-B/16~\cite{dosovitskiy2021an} as our image encoder with the input image resolution of 384$\times$384. Overall, \emph{only base text and image encoders} are considered in our COTS.

\noindent\textbf{Evaluation Metrics.}~~The widely-used R@$k$ ($k=1,5,10$) in cross-modal retrieval is reported for performance evaluation, which is the proportion of matched samples found in the top-$k$ retrieved results. Following~\cite{bain2021frozen}, we also report the Median Rank (MR) for video-text retrieval.

\noindent\textbf{Implementation Details.}~~For our masked vision-language modeling (MVLM), we randomly mask 40\% image patches following~\cite{bao2021beit} and mask word tokens in text with 15\% probability. 
We adopt the Adam~\cite{kingma2014adam} optimizer with a weight decay of 0.02. We select hyper-parameters heuristically due to computational constraint: the momentum hyper-parameter $m = 0.99$, temperature $\tau = 0.05$, and the queue size $N_Q$ is 12,800, 6,400, and 1,200 for pre-training, finetuning on MSCOCO, and finetuning on Flickr30K, respectively. 
We set the initial learning rate to 5e-5 for the first 5 epochs, and decay the learning rate linearly in the rest epochs.
More implementation details can be found in the supp. material.

\vspace{-0.1cm}
\subsection{Image-Text Retrieval}
\label{sec:main_results}
\vspace{-0.1cm}

\noindent\textbf{Comparison to the State-of-the-Arts.}~~We compare our COTS with the state-of-the-art methods on two widely-used image-text datasets: Flickr30K and MSCOCO. As shown in Table~\ref{tab:image_sota}, under a fair comparison setting (excluding VSE$\infty^{*\dagger}$~\cite{chen2021learning} and COOKIE$^{*\dagger}$~\cite{wen2021cookie} which utilize 940M tagged images for visual encoder pre-training), our COTS outperforms all two-stream models by large margins for all evaluation metrics. Specifically, compared with the most recent two-stream model COOKIE~\cite{wen2021cookie}, our COTS achieves higher results by 5.2\%~(66.9\% vs. 61.7\%) for I2T R@1 on MSCOCO and 3.9\%~(50.5\% vs. 46.6\%) for T2I R@1 on MSCOCO, but with less pre-training data~(5.3M vs. 5.9M).
Moreover, when leveraging larger pre-training dataset and model ensemble technique, our COTS$^\dagger$ further improves the performance. Specifically, without using extra object detectors, our COTS$^\dagger$ achieves new state-of-the-art on Flickr30K w.r.t. both single-stream and two-stream methods. On MSCOCO, our COTS$^\dagger$ also achieves higher performance than most single-stream methods and comparable results compared with VinVL~\cite{zhang2021vinvl} but with a 10,800$\times$ faster speed during inference (see Inference Efficiency Analysis).

\begin{figure}[t]
    \centering
    \includegraphics[width=0.96\linewidth]{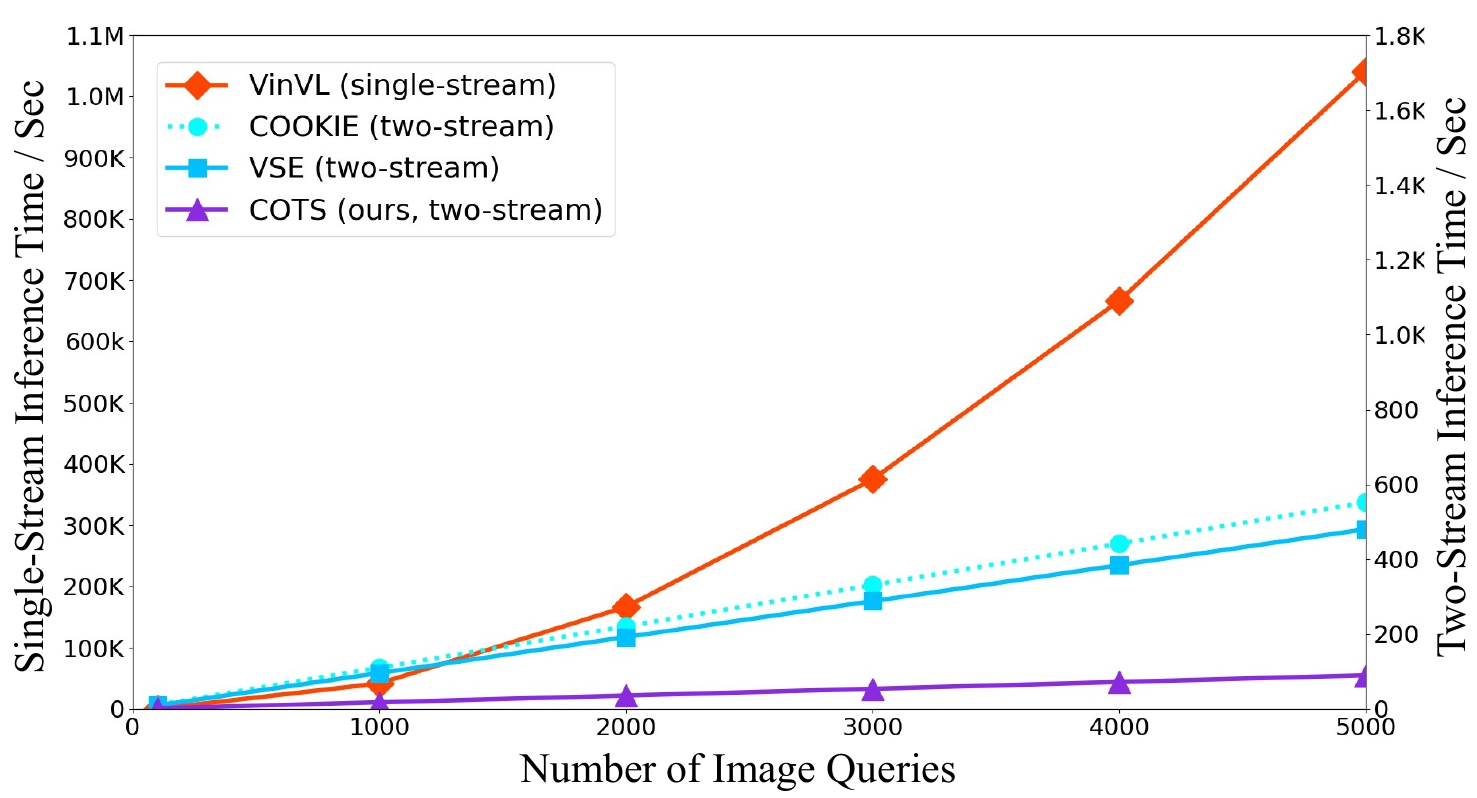}
    \vspace{-0.11in}
    \caption{
    Comparison of inference time to different methods (VinVL~\cite{zhang2021vinvl}, COOKIE~\cite{wen2021cookie}, VSE~\cite{chen2021learning}) on MSCOCO (5K) test set.
    }
    \label{fig:efficiency}
    \vspace{-0.09in}
\end{figure}

\noindent\textbf{Inference Efficiency Analysis.}~~In real-world application scenarios, inference speed is an important evaluation metric for retrieval methods. In Figure~\ref{fig:efficiency}, we compare our COTS with recent state-of-the-arts regarding the inference time on the MSCOCO~(5K) test set. 
All methods are evaluated on a single Tesla V100 GPU. 
Compared with the single-stream VinVL~\cite{zhang2021vinvl}, our COTS is 10,800$\times$ faster on the whole MSCOCOC~(5K) test set. This huge gap will even become dramatically larger when the size of test set $N$ grows, as the retrieval time complexity for single-stream models is $O(N^2)$ while it is nearly $O(N)$ for two-stream ones. Although VSE and COOKIE are also two-stream models, our COTS is still significantly faster than them, indicating the extreme high efficiency of our COTS due to its fully tokenized Transformer-based architecture.

\begin{table}[t]
    \centering
    \scalebox{0.85}{
    \tabcolsep5.4pt
    \begin{tabular}{lcccccc}
    \toprule
    \multirow{2}{*}{Model}  & \multicolumn{3}{c}{I2T Retrieval} & \multicolumn{3}{c}{T2I Retrieval} \\
     & R@1 & R@5 & R@10 & R@1 & R@5 & R@10\\
    \midrule
    ViLT~\cite{kim2021vilt} & 56.5 & 82.6 & 89.6 & 40.4 & 70.0 & 81.1  \\
    CLIP~\cite{radford2021learning} & 58.4 & 81.5 & 88.1 & 37.8 & 62.4 & 72.2  \\
    ALIGN~\cite{jia2021scaling} & 58.6 & 83.0 & 87.9 & \bf45.6 & 69.8 & 78.6  \\
    \midrule
    COTS~(w/o FT) & \bf60.4 & \bf84.7 & \bf91.7 & 43.8 & \bf71.6 & \bf81.3\\
    \bottomrule
    \end{tabular}}
    \vspace{-0.1in}
    \caption{Comparative results (without finetuning) for image-text retrieval on the MSCOCO (5K) test set. FT -- Finetuning.}
    \label{tab:zeroshot}
    \vspace{-0.2in}
\end{table}

\noindent\textbf{Comparative Retrieval Results without Finetuning.}~~Following ViLT~\cite{kim2021vilt}, we report the comparative retrieval results without finetuning on MSCOCO in Table~\ref{tab:zeroshot}. We can observe that: (1) Our COTS outperforms the latest single-stream method ViLT~\cite{kim2021vilt}. (2) Our COTS also beats the latest two-stream methods CLIP~\cite{radford2021learning} and ALIGN~\cite{jia2021scaling}, although it is pre-trained with much less data.

\begin{table}[t]
    \centering
    \scalebox{0.85}{
    \tabcolsep4pt
    \begin{tabular}{lcccccc}
    \toprule
\multirow{2}{*}{Method}  & \multicolumn{3}{c}{I2T Retrieval} & \multicolumn{3}{c}{T2I Retrieval} \\
     & R@1 & R@5 & R@10 & R@1 & R@5 & R@10\\
    \midrule
    $\mathcal{L}_{\text{inst}}$ & 24.0 & 48.3 & 60.0 & 16.8 & 37.5 & 49.6  \\
    $\mathcal{L}_{\text{inst}} + \mathcal{L}_{\text{CMLM}}$ & 24.5 & 49.3 & 61.1 & 16.5 & 37.8 & 49.9  \\
    $\mathcal{L}_{\text{inst}} + \mathcal{L}_{\text{token}}$ & 25.6 & 49.9 & 61.9 & 17.1 & 38.3 & 50.4  \\
    $\mathcal{L}_{\text{inst}} + \mathcal{L}_{\text{token}} + \mathcal{L}_{\text{task}}$ & 26.4 & 50.5 & \bf62.9 & 17.5 & 38.5 & 50.6  \\
    \midrule
    $\mathcal{L}_{\text{inst}}$ (w/ AMF) & 24.7 & 49.6 & 61.3 & 16.6 & 38.3 & 50.0\\
    Our Full COTS & \bf27.1 & \bf51.1 & \bf62.9 & \bf17.9 & \bf39.2 & \bf51.1\\
    \bottomrule
    \end{tabular}}
    \vspace{-0.1in}
    \caption{Ablation study for our COTS pre-trained on the small CC200K dataset. Zero-shot image-text retrieval results are reported on the MSCOCO (5K) test set. }
    \label{tab:ablation}
    \vspace{-0.1in}
\end{table}

\begin{table}[t]
    \centering
    \scalebox{0.85}{
    \tabcolsep5pt
    \begin{tabular}{lrcccc} 
    \toprule
    Model & \# PT Pairs & R@1 & R@5 & R@10 & MR$\downarrow$  \\ 
    \midrule
    \bf{7K split:}  & \\
    JSFusion~\cite{yu2018joint}  & - & 10.2 & 31.2 & 43.2 & 13.0   \\
    HT MIL-NCE~\cite{miech2019howto100m} & $>$100M & 14.9 & 40.2 & 52.8 & 9.0   \\
    ActBERT~\cite{zhu2020actbert}  & $>$100M & 16.3 & 42.8 & 56.9 & 10.0  \\
    HERO~\cite{li2020hero}  & $>$100M & 16.8 & 43.4 & 57.7 & -  \\
    VidTranslate~\cite{korbar2020video}  & $>$100M & 14.7 & - & 52.8  &  - \\
    NoiseEstimation$^*$~\cite{amrani2020noise}  & $>$100M & 17.4 & 41.6 &  53.6 & 8.0 \\
    UniVL$^*$~\cite{luo2020univilm} & $>$100M & 21.2 & 49.6 & 63.1 & 6.0  \\ 
    ClipBERT~\cite{lei2021less}  & 5.6M & 22.0 & 46.8 & 59.9 & 6.0   \\
    TACo$^*$~\cite{yang2021taco} & $>$100M & 24.8 & 52.1 & 64.5 & 5.0 \\
    COTS~(ours)  & 5.3M & 29.0 & 57.0 & 67.7 & 3.0 \\
    COTS~(ours)  & 15.3M & \bf32.1 & \bf60.8 & \bf70.2 & \bf3.0 \\
    \midrule
    \bf{1KA split:}  & \\
    AVLnet$^*$~\cite{rouditchenko2020avlnet}  & $>$100M & 27.1 & 55.6 & 66.6 & 4.0  \\
    MMT$^*$~\cite{gabeur2020multi} & $>$100M & 26.6 & 57.1 & 69.6 & 4.0 \\
    TACo$^*$~\cite{yang2021taco} & $>$100M & 28.4 & 57.8 & 71.2 & 4.0 \\
    Support Set$^*$~\cite{patrick2020support}  & $>$100M & 30.1 & 58.5 &69.3 & 3.0  \\
    Frozen in Time~\cite{bain2021frozen}  & 5.5M & 31.0 & 59.5 & 70.5 & 3.0 \\ 
    COTS~(ours)  & 5.3M & 33.1 & 61.3 & 72.8 & 3.0 \\
    COTS~(ours)  & 15.3M & \bf36.8 & \bf63.8 & \bf73.2 & \bf2.0 \\
    \bottomrule
    \end{tabular}
    }
    \vspace{-0.1in}
    \caption{Comparison to the state-of-the-arts for text-to-video retrieval on MSR-VTT under two splits: the 7K and 1KA splits. Notations: $\downarrow$ denotes that lower results are better; $^*$ denotes that extra modalities (e.g., motion and audio) are used. }
    \label{tab:video-sota}
    \vspace{-0.25in}
\end{table}

\begin{figure*}[ht]
    \centering
    \includegraphics[width=0.99\textwidth]{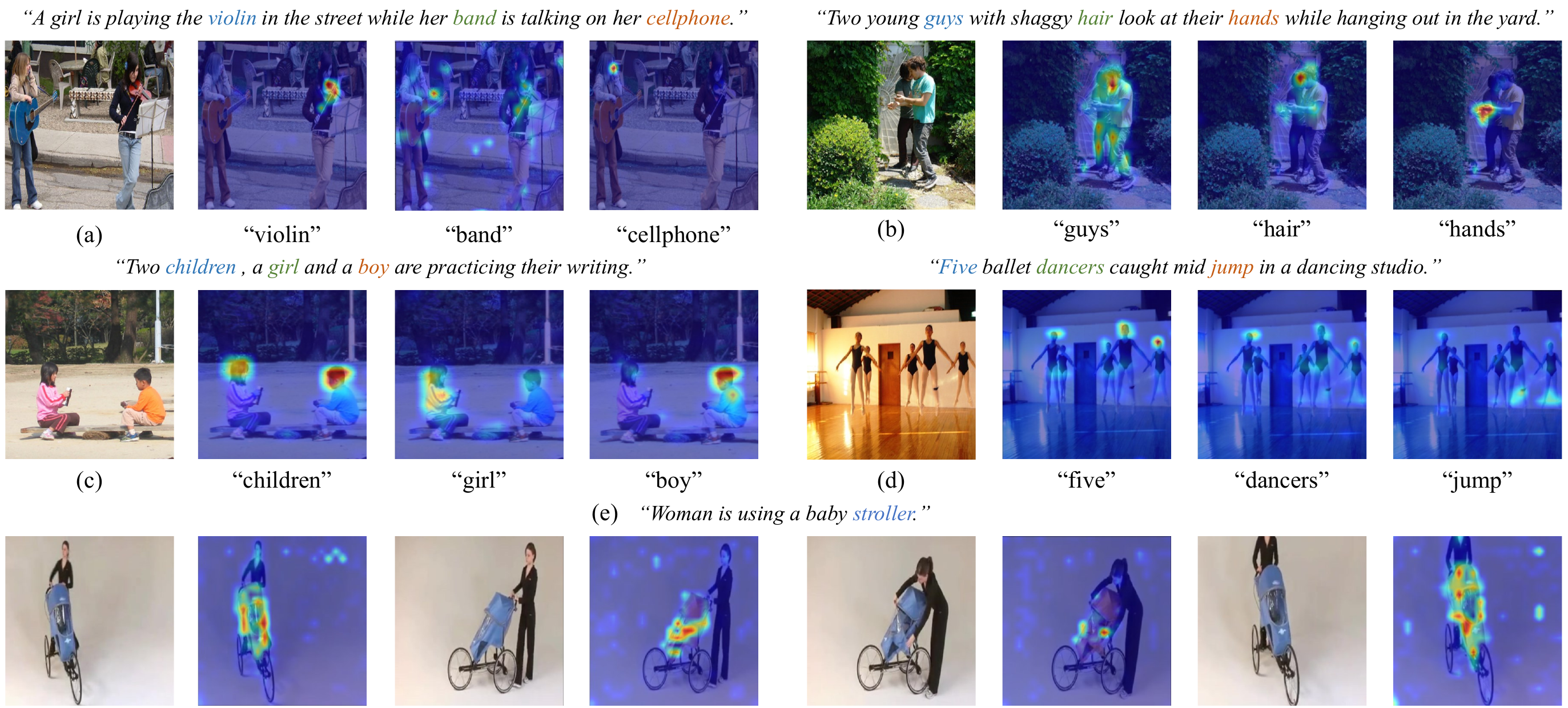}
    \vspace{-0.1in}
    \caption{
    Visualizations of attention maps of our COTS using GAE~\cite{chefer2021generic} on images/video frames responding to individual words. (a) -- (d) Image attention maps w.r.t. different words. (e) Video frame attention maps w.r.t. the word ``stroller''.
    }
    \label{fig:vis}
    \vspace{-0.1in}
\end{figure*}

\noindent\textbf{Ablation Study Results.}~~In Table~\ref{tab:ablation}, we analyze the contributions of different pre-training objectives and the adaptive momentum filter (AMF) module in our COTS. We randomly sample 200K image-text pairs from CC12M as the pre-training dataset (termed CC200K). Zero-shot retrieval results are reported on the MSCOCO (5K) test set. We start with our instance-level interaction loss $\mathcal{L}_{\text{inst}}$ (without AMF) and then add other losses successively. We can observe from Table~\ref{tab:ablation} that: (1) Both CMLM and CMVM bring performance improvements (see $\mathcal{L}_{\text{inst}} + \mathcal{L}_{\text{CMLM}}$ vs. $\mathcal{L}_{\text{inst}}$, and $\mathcal{L}_{\text{inst}} + \mathcal{L}_{\text{token}}$ vs. $\mathcal{L}_{\text{inst}} + \mathcal{L}_{\text{CMLM}}$), indicating that token-level cross-modal interactions are beneficial to learning the aligned multi-modal representation space. (2) When task-level interaction is added (see $\mathcal{L}_{\text{inst}} + \mathcal{L}_{\text{token}} + \mathcal{L}_{\text{task}}$ vs. $\mathcal{L}_{\text{inst}} + \mathcal{L}_{\text{token}}$), the performance is further improved, which clearly validates the effectiveness of our multi-level cross-modal interactions. (3)~Our AMF module works well with either instance-level or multi-level interactions (see $\mathcal{L}_{\text{inst}}$ (w/ AMF) vs. $\mathcal{L}_{\text{inst}}$, and Our Full COTS vs. $\mathcal{L}_{\text{inst}} + \mathcal{L}_{\text{token}} + \mathcal{L}_{\text{task}}$). (4) Combining all objectives with the AMF module (i.e., Our Full COTS) leads to the best results, indicating that each objective/module is complementary to each other.

\subsection{Video-Text Retrieval}
\label{video}

We further compare our COTS with the state-of-the-art methods on the video-text retrieval task. To directly deploy our COTS, we do not consider utilizing complex methods or additional modules to model the temporal information of videos. Instead, we simply use the mean frame embeddings as video representations and then calculate similarity scores by dot product with text embeddings.
We report the text-to-video retrieval results on the MSR-VTT dataset in Table~\ref{tab:video-sota}. Note that only text-to-video retrieval is considered as in the latest work~\cite{bain2021frozen}. It can be seen that: (1) Our COTS significantly outperforms the state-of-the-arts even without modeling the temporal information of videos, which demonstrates the general applicability and the great potentiality of our COTS. (2) Our COTS leads to better results than methods utilizing extra modalities (e.g., motion and audio) or those pre-trained on extra large video data (e.g., the HowTo100M dataset~\cite{miech2019howto100m} with more than 100 million video-text pairs), indicating that a well pre-trained vision-language model may be the key to video-text retrieval.

\subsection{Visualization Results}

Figure~\ref{fig:vis} shows the visualized attention maps of our COTS on images/video frames responding to individual words. We can see from Figures~\ref{fig:vis}(a)--(b) that our COTS can well locate different objects (even fine-grained ones like ``violin'' and ``cellphone'' in Figure~\ref{fig:vis}(a), ``hair'' and ``hands'' in Figure~\ref{fig:vis}(b)) in the same image. Figure~\ref{fig:vis}(c) shows how our COTS determines gender information. Given the word ``children'', COTS focuses on the faces. When recognizing ``girl'', COTS pays attention to the girl's long hair and pink clothes (and the same for the word ``boy''). Interestingly, our COTS can also capture abstract concepts (``five'') and actions (``jump'') as shown in Figure~\ref{fig:vis}(d). COTS focuses on five dancers for both ``five'' and ``dancers'', but pays more attention for the number ``five''. And it focuses on feet when it comes to ``jump''. Figure~\ref{fig:vis}(e) presents attention maps w.r.t. ``stroller'' on four frames from the same video, showing that our COTS can also work well for the video modality.

\section{Conclusion}
 
In this paper, we have investigated how to improve the performance of the two-stream vision-language pre-training (VLP) while still maintaining its advantage of high efficiency for image-text retrieval. Specifically, we propose a novel COllaborative Two-Stream VLP model termed COTS by leveraging three levels of cross-modal interactions in image-text retrieval. That is, we consider token-level interaction by masked vision-language modeling with both tokenized images and texts, instance-level interaction by cross-modal momentum contrastive learning, and task-level interaction by aligning two task distributions. Extensive experiments validate the effectiveness and high efficiency of our COTS in image-text retrieval. It is also shown to have general applicability as it achieves new state-of-the-art on video-text retrieval without any modification.

{
\noindent\textbf{Acknowledgements}~~~This work was supported in part by National Natural Science Foundation of China (61976220 and 61832017), Beijing Outstanding Young Scientist Program (BJJWZYJH012019100020098), and Large-Scale Pre-Training Program 468 of BAAI.
}

{\small
\bibliographystyle{ieee_fullname}
\bibliography{COTS-VLM}
}

\end{document}